\documentclass[a4paper]{svproc}
\usepackage{url}
\usepackage{graphicx}
\usepackage{booktabs}
\usepackage{subfig}
\usepackage{xcolor}
\usepackage{amsmath,amssymb,amsfonts}
\usepackage{bbold}
\usepackage{relsize}

\usepackage[sort,compress]{cite}
\usepackage[linesnumbered,ruled,vlined]{algorithm2e}
\SetKwInput{KwInput}{Input}
\SetKwInput{KwOutput}{Output}

\graphicspath{{figures/}}

\newcommand*\ie{i.e.}  % i.e.
\newcommand*\etal{\emph{et al.}}  % et al.

\newcommand*\mx[2][{}]{{}^{#1}\tens{#2}} % Matrix command
\newcommand*\vc[2][{}]{{}^{#1}\vec{#2}} % Vector command
\newcommand*\manX{X}  % manifold X
\newcommand*\setXsafe{\manX_{\text{safe}}}  % Xsafe
\newcommand*\setU{\mathcal{U}}  % control set
\newcommand*\R{\mathbb{R}}  % real space
\newcommand*\ttt[1]{\texttt{#1}} % texttt
\newcommand*\T{\ensuremath{\top}}  % transpose

\newcommand*{\rot}[1][{\,}]{{}^{{#1}\!}{q}}  % rotation matrix
\newcommand*{\rothat}[1][{\,}]{{}^{{#1}\!}{\hat q}}  % rotation matrix
\newcommand*\mxhat[2][{\,}]{{}^{#1\!}\hat{{#2}}} % Matrix command (with dot for time derivative)
\newcommand*\vchat[2][{\,}]{{}^{#1\!}\hat{\vc{#2}}} % Vector command (with dot for time derivative)
\newcommand*\vctilde[2][{\,}]{{}^{#1\!}\tilde{\vc{#2}}} % Vector command (with dot for time derivative)

\newcommand*\diff{\ensuremath{\mathop{}\!\mathrm{d}}}  % diff operator
\newcommand*\ddt{\ensuremath{\mathop{}\!\frac{\diff}{\diff t}}}  % liebniz d/dt operator
\newcommand*\tr{\ensuremath{\mathrm{tr}\mathop{}\!}}  % trace
\newcommand*\blkdiag{\ensuremath{\texttt{blkdiag}\!}}  % block diagonal
\newcommand*\E{\ensuremath{\mathbb{E}\mathop{}\!}}  % expectation
\renewcommand*\skew[1]{\ensuremath{[#1]_{\times}}}  % skew-symm matrix
\newcommand*\kron{\otimes}
\newcommand*\rflex{{r_0}}

\newcommand*\Lam{\mx{\Lambda}}
\newcommand*\LamZ{\Lam^{\mathcal{Z}}}
\newcommand*\Ljh[1][j]{{}^{{#1}\!}L_h}
\newcommand*\Hlj[1][j]{{}^{{#1}\!}\mx{H}}
\newcommand*\Hlji[1][j]{{}^{{#1}\!}_{i\!\!\!}\mx{H}}
\newcommand*\Llj[1][j]{{}^{{#1}\!}\mx{L}}
\newcommand*\Hljn[1][j]{\Hlj^{(n)}}
\newcommand*\Lljn[1][j]{\Llj^{(n)}}

\newcommand*\elln[1][n]{\ell^{(#1)}}
\newcommand*\lmn[1][n]{\vc{l}^{(#1)}}
\newcommand*\dlmn[1][n]{{\vc{\delta l}^{(#1)}}}  % differential landmark l
\newcommand*\Dt{\Delta t}
\renewcommand*\P{\mx{P}}
\renewcommand*\S{\mx{S}}
\newcommand*\Jobs{J_\text{obs}} % J_observability
\newcommand*\vizind{\mathbb{1}_\text{vis}} % viz indicator
\newcommand*\Kbar{{\bar{K}}}  % number of intervals
\newcommand*\kbar{{\bar{k}}}  % interval iterator
\newcommand*\LmDist{\mathcal{M}}  % abstract landmark distribution

\newcommand*\dx{{\vc{\delta x}}}  % differential state x
\newcommand*\du{{\vc{\delta u}}}  % differential control u
\newcommand*\dzn[1][n]{{\vc{\delta}\zn[#1]}}  % differential observation z
\newcommand*\e{\vc{e}}  % estimation error
\newcommand*\w{\vc{w}}  % process noise
\newcommand*\MVN{\mathcal{N}}  % Normal distribution symbol

\newcommand*\xhat{\vc{\hat{x}}}

\newcommand*\lmntilde[1][n]{\vc{\tilde{l}}^{({#1})}}
\newcommand*\zn[1][n]{{\vec{z}^{(#1)}}}

\newcommand*\nun[1][n]{{\vec{\nu}^{(#1)}}}
\newcommand*\zntildek[1][k]{{\vec{\tilde{z}}_{#1}^{(n)}}}
\newcommand*\nunk[1][k]{{\vec{\nu}_{#1}^{(n)}}}

\newcommand*\Ablock{\mathcal{A}}

\newcommand*\Gblock{\mathcal{G}}
\newcommand*\W{\mathcal{W}}
\newcommand*\Hblock[1][n]{\mathcal{H}}
\newcommand*\Hbarblock[1][n]{\overline{\mathcal{H}}}
\newcommand*\Hibarblock[1][i]{{}_{#1\!}\Hbarblock}
\newcommand*\Hbarblockexp[1][n]{\Hbarblock_{\text{exp}}}
\newcommand*\Hbarblocklie[1][n]{\Hbarblock_{\text{lie}}}
\newcommand*\Lblock[1][n]{\mathcal{L}}
\newcommand*\Hnblock[1][n]{\mathcal{H}^{({#1})}}
\newcommand*\Lnblock[1][n]{\mathcal{L}^{({#1})}}
\newcommand*\Nun[1][n]{\mathcal{V}^{({#1})}}
\newcommand*\Qn[1][n]{{\mathcal{Q}^{({#1})}}}
\newcommand*\Zn[1][n]{\mathcal{Z}^{({#1})}}
\newcommand*\Znexp[1][n]{\Zn[#1]_\text{exp}}
\newcommand*\Znlie[1][n]{\Zn[#1]_\text{lie}}
\newcommand*\blockM{\mx{M}}

\newcommand*\pT{\mathcal{T}}

\newcommand{\xx}[1]{{\color{red} {\bf XX #1 XX\ }}}
\newcommand{\XX}[1]{{\color{magenta} {\bf XX #1 XX\ }}}

\renewcommand{\xx}[1]{}
\renewcommand{\XX}[1]{}

\begin{document}
\mainmatter              % start of a contribution
\title{Towards Online Observability-Aware Trajectory Optimization for Landmark-based Estimators}
\subtitle{Technical Report}
\titlerunning{Observability-Aware Traj.\ Optimization for Landmark-based Estimators}  % abbreviated title (for running head)
%\toctitle{Observability-Aware Trajectory Optimization for Landmark-based Estimators}
\author{Kristoffer M. Frey\inst{1,2} \and Ted Steiner\inst{2}
\and Jonathan P. How\inst{1}}
\authorrunning{Kristoffer M. Frey et al.} % abbreviated author list (for running head)
%\tocauthor{Kristoffer Frey, Ted Steiner, Jonathan P. How}
%
\institute{
  Massachusetts Institute of Technology, Cambridge MA 02139, USA,\\
  \email{kfrey,jhow@mit.edu}
  \and
  The Charles Stark Draper Laboratory, Inc., Cambridge MA 02139, USA,\\
  \email{tsteiner@draper.com}
}

\maketitle              % typeset the title of the contribution

\begin{abstract}
As autonomous systems increasingly rely  on onboard sensing for localization and perception, the  parallel tasks of motion planning and state estimation become more strongly coupled.
This coupling is well-captured by augmenting the planning objective with a posterior-covariance penalty -- however, prediction of the estimator covariance is challenging when the observation model depends on unknown landmarks, as is the case in Simultaneous Localization and Mapping (SLAM).
This paper addresses these challenges in the case of landmark- and SLAM-based estimators, enabling efficient prediction (and ultimately minimization) of this performance metric.
First, we provide an interval-based filtering approximation of the SLAM inference process which allows for recursive propagation of the ego-covariance while avoiding the quadratic complexity of explicitly tracking landmark uncertainty.
Secondly, we introduce a Lie-derivative measurement bundling scheme that simplifies the recursive ``bundled'' update, representing significant computational savings for high-rate sensors such as cameras.
Finally, we identify a large class of measurement models (which includes orthographic camera projection) for which the contributions from each landmark can be directly combined, making evaluation of the information gained at each timestep (nearly) independent of the number of landmarks.
This also enables the generalization from finite sets of landmarks $\{\elln\}$ to \emph{distributions}, foregoing the need for fully-specified linearization points at planning time and allowing for new landmarks to be anticipated.
Taken together, these contributions allow SLAM performance to be accurately and efficiently predicted, paving the way for online, observability-aware trajectory optimization in unknown space.

% We would like to encourage you to list your keywords within
% the abstract section using the \keywords{...} command.
  % TODO: any more?
\end{abstract}
\section{Introduction}
  In the last decade, significant progress has been made to enable basic autonomy for low-SWaP (Size, Weight, and Power) systems.
  Thanks to recent algorithmic advances, such systems can navigate purely from onboard sensors and avoid dependence on dedicated infrastructure such as GPS or motion-capture, allowing operation in a wider range of environments.
  However, commonly-used sensors such as IMUs, laser scanners, and cameras have nonlinear observation models and/or limited field-of-view (FoV), and therefore the observability of the estimated state will fundamentally depend on the system trajectory.
  Furthermore, these sensors can have time-varying latent parameters (e.g.,\ IMU biases, rigid-body calibrations) and environmental dependencies (e.g.,\ the presence or absence of high-gradient corner features).
  Thus, even with a good initialization point, estimation quality can degrade to catastrophic levels if the chosen trajectory or environment does not provide sufficient information.

  In this paper, we are interested in discovering well-observable motions accounting for the (possibly latent) distribution of landmarks over receding time-horizons.
  This motivates continuous optimization techniques leveraging a full-DoF system model, and efficient evaluation requires a compact representation of the SLAM inference process.
  In focusing on receding time-horizons, we are willing to neglect certain aspects of SLAM such as global loop closure in favor of computational efficiency and applicability to real-world scenarios where the map is not fully known at planning time.

  %Conventional planning approaches~\cite{mellnnger2011minimum} minimize energy or control effort along a trajectory, an objective often in tension with observability.
  %However, maintaining calibrations for a fleet of vehicles can be onerous, and in many environments the distribution of landmarks is far from uniform.
  %In these cases, designing trajectories that ensure good observability can produce significant localization improvement in general and may be the difference between mission success and failure.

  When formulating the uncertainty-aware planning problem, we follow the common approach~\cite{prentice2009belief,bry2011rapidly,van2011lqg,van2012motion,indelman2015planning,zhang2018perception} of representing uncertainty as a Gaussian distribution about a nominal trajectory.
  This representation is deterministic and compactly parameterizable, and is therefore amenable to optimization based on trajectory sampling~\cite{prentice2009belief,bry2011rapidly,van2011lqg}, motion-primitives~\cite{elisha2017active,kopitkov2017no,elimelech2017scalable,zhang2018perception}, or continuous optimization~\cite{indelman2015planning,van2012motion}.
  Of these, sampling-based approaches are not well-suited for systems with more than a few degrees of freedom, and motion-primitive solutions do not address the fundamental issue of ensuring trajectories are well-observable in the first place.
  In the space of continuous trajectory optimization, existing approaches have only been demonstrated with relatively simple sensor models~\cite{preiss2018simultaneous,rafieisakhaei2017use,van2012motion} or scale poorly to large numbers of landmarks~\cite{indelman2015planning}.

  In order to avoid the computational challenges associated with explicit covariance prediction and minimization, heuristic approaches maximize an observability or (in the case of landmark-based systems) visibility metric.
  For example,~\cite{conticelli2000observability,mariottini2009vision,arneberg2018guidance} identify and enumerate non-observable configurations to hand-design a discrete set of ``well-observable'' maneuvers for use in online planning.
  More recently,~\cite{nobre2019learning} leverages a reinforcement learning framework to select a sequence of primitive maneuvers in a manual calibration routine.
  Working from a continuous optimization framework, \cite{krener2009measures,preiss2018simultaneous} design trajectories which maximize observability metrics based on the Local Observability Gramian (LOG).
  For landmark-based systems with limited FoV (e.g.,\ visual SLAM), explicit visibility-based metrics have been used by~\cite{falanga2018pampc} and~\cite{murali2019perception} in real-time planning frameworks, with some success.
  However, LOG and visibility objectives are ultimately heuristic and therefore may select high-energy (expensive) trajectories with little corresponding estimation improvement~\cite{rafieisakhaei2017use}.

  Nevertheless, direct posterior-covariance minimization is challenging, particularly in the case of SLAM.
  Landmark-based systems such as visual SLAM can estimate tens or hundreds of landmarks simultaneously, and linearization points of to-be-discovered landmarks are generally unknown at planning time.
  Besides having to compute and accumulate the contribution from each of these landmarks (linear complexity), explicit representation of the belief state requires either reasoning over random factor graphs or maintaining the joint covariance matrix over the ego-state \emph{and} map, both of which are intractable in online application.
  Furthermore, high-rate sensors such as cameras can produce many measurements over the time-horizon in consideration, furthermore contributing to the complexity of direct covariance analysis.
  While~\cite{kopitkov2017no} and~\cite{elimelech2017scalable} introduce some sparsification and incremental-update techniques to partially mitigate the cost of graph-based evaluation with large maps, they require known landmark linearization points and are applicable only in the case of finite action sets.

  \subsection{Contributions}
  In this work, we address these challenges to enable efficient posterior-covariance prediction and minimization for a general class of landmark-based observation models.

  \begin{itemize}
    \item \textbf{Measurement bundling:}
    To avoid the prohibitive complexity of forward-simulating the SLAM estimation process during motion planning, Section \ref{s:interval_filter} describes a compact, filtering-based approximation that avoids representation of the landmark uncertainty entirely.
    We refer to this simplified framework as the Structureless Interval Information Filter (SIIF), as it ``bundles'' measurements according to pre-specified time intervals and marginalizes out the uncertain map state from each ``interval update.''

    \item \textbf{A Lie-Taylor update approximation:}
    The exact or ``explicit'' form of the SIIF update represents the full history of discrete-time process noise within the interval and requires multiple rounds of state integration, resulting in non-trivial computation.
    To simplify implementation and reduce computation, Section \ref{s:lie_taylor_update} introduces an approximate form of the interval update based on continuous-time Lie derivatives.
    This provides significant computational speedups (particularly in the case of high-rate sensors like cameras) while preserving the observability characteristics of the system.

    \item \textbf{Handling large numbers of unknown landmarks:}
    In general, the information contribution from each landmark must be computed individually, rendering the SIIF update expensive when landmarks number in the hundreds and ill-posed when linearization points are unavailable.
    Fortunately, we identify a convenient (and useful) class of sensor models for which the total information against a landmark \emph{distribution} can be computed directly.
    From a purely computational perspective, this facilitates scaling to large clouds of hundreds or even thousands of landmarks, and ultimately allows planning against predictive models of feature density in the environment.
    As compared to~\cite{zhang2019beyond}, which only applies to a particular observation and visibility model of \emph{known} landmarks, our decomposition applies to a much more general class of models which includes orthographic camera projection.
  \end{itemize}

  In Section \ref{s:results} we numerically validate our measurement bundling approximation, demonstrating significant computational improvement with low approximation error.
  Moreover, we evaluate our full trajectory generation pipeline over a large number of random trials, demonstrating more effective estimation improvement than heuristic methods (in a Pareto sense).
  Although a fully online planning implementation is still in-work, the improvements and results herein suggest that real-time performance should be attainable.

\section{Preliminaries}
  We assume the system state $x$ lies on an $n_x$-dimensional manifold $\manX$ with dynamics given by the stochastic ODE
  \begin{equation}\label{eq:dyn}
    \diff x(t) = f\big(x(t), u(t) \big) \diff t + \mx{G}_c\big(x(t), u(t) \big) \diff \w(t).
  \end{equation}
  The input $u(t)$ is confined to a set $\setU$, and $\vc{w}(\cdot)$ is a standard Brownian noise process.
  %Besides the dynamic state $x$, the system and sensors may be characterized by a set of (assumed constant) parameters $\theta$, which may include uncertain quantities like IMU biases, aerodynamic coefficients, camera calibrations, etc.
  %Though in the estimation literature \cite{bloesch2017iterated,steiner2017vision} these parameters are often treated as slowly time-varying, this simplifying assumption is adequate for the purposes of short- to medium-horizon planning and greatly reduces algorithmic complexity.
%
  We assume measurements are acquired by a collection of homogeneous sensors
  \begin{equation}\label{eq:obs}
    \zn = h(x; \elln) + \nun \quad \in \R^m,
  \end{equation}
  each corrupted by an independent Gaussian noise $\nun$ and parameterized by a latent environmental variable $\elln$ (i.e.,\ a \emph{landmark}).
  We assume that these measurements are acquired synchronously with frequency $1/\Dt$.
  Without loss of generality, each $\nun$ is assumed to have identity covariance.

  We are interested in the setting where state $x(t)$ and landmarks $\{ \elln \}$ are all uncertain, and that the system controller must operate from an \emph{estimated} state $\xhat(t)$.
  It is worth noting that our framework avoids making any implementation-specific assumptions with regard to this ``estimator,'' aside from the fact that it is constrained to operate as a function of the observation history and initializes from a Gaussian prior with mean $\xhat_0$ and covariance $\P_0$.
  For the purposes of planning well-observable trajectories, this paper will apply a local Linear-Time-Varying (LTV) assumption which simplifies computation of the ``optimal'' error covariance as a deterministic function of the underlying trajectory.
  It is worth emphasizing that under this framework, the posterior error covariance $\P(T)$ reflects the \emph{intrinsic} (local) observability properties of the trajectory in implementation-agnostic way.

  \subsection{Local Linear-Time-Varying (LTV) dynamics}
    Let $\big(x(\cdot), u(\cdot) \big)$ define a nominal (noise-free) state/control trajectory pair, that is it obeys $\dot{x} = f(x, u)$.
    The small-perturbation dynamics can then be approximated by linearizing $f$ and $h$ about $(x,u)$ and the landmark linearization points $\{\elln\}$.
    This gives a stochastic LTV system
    \begin{subequations}\label{eq:ltv}
      \begin{align}
        \diff \dx(t) &\approx \Big( \mx{A}_c(t) \dx(t) + \mx{B}_c(t) \du(t) \Big)\diff t + \mx{G}_c(t) \diff \vc{w}(t)  \label{eq:dyn_ltv}  \\
        \dzn(t) &\approx \mx{H}\big( x(t); \elln \big) \dx(t) + \mx{L}\big( x(t); \elln \big) \dlmn + \nun(t)  \label{eq:obs_ltv}
      \end{align}
    \end{subequations}
    where we have the dropped the explicit dependence on the linearization point $\big( x(t), u(t), \elln \big)$ for clarity.
    Note that the local observation model (\ref{eq:obs_ltv}) depends not only on the state deviation $\dx(t)$ but also the deviation of the landmark $\dlmn$.

    The estimator error covariance $\P(t)$ is defined with respect to an assumed error state $\e(t) = \xhat(t) \boxminus x(t)$, where the $\boxminus$ operator reflects a generalized ``difference'' operator on the manifold $\manX$.
    Crucially, for reasons that will be elaborated later, we specifically avoid representing the landmark deviations in $\e(t)$, which would have quadratic complexity implications on the matrix $\P(t)$.
    This will be accomplished via a marginalization approach similar in spirit to the MSCKF \cite{mourikis2007multi}, and is motivated by the fact that we seek an efficient proxy for estimator performance (capturing ``observability'' as a function of a trajectory) rather than an actual estimator implementation.

  \subsection{Trajectory Optimization}
    This paper focuses on deterministic, continuous optimal control problems of the form
    \begin{alignat}{3}
      \min_{u(\cdot)}& \qquad J_c(u) + \lambda \Jobs(u)  \label{pblm:ct}  \\
      \text{subject to:}& \qquad \dot{x}(t) = f\big(x(t), u(t) \big)  \qquad  &&x(0) = x_0  \label{pblm:dynamics}  \\
                        & \qquad u(t) \in \setU                      \qquad  &&x(t) \in \setXsafe  \nonumber
    \end{alignat}
    where we have augmented a ``conventional'' cost functional $J_c$, representing a min-energy or min-time objective, with auxiliary penalty $\Jobs$ capturing the effects of uncertainty.
    Note that the dynamics constraint (\ref{pblm:dynamics}) is defined via a noise-free, nominal model.

    Note that irrespective of the choice of $\Jobs$, the presence of nonlinear dynamics is generally sufficient to imply that only numeric, local solutions are available in practice.
    Thus, choice of a non-convex $\Jobs$ does not usually make (\ref{pblm:ct}) fundamentally harder to solve.
    However, it is critical that $\Jobs$ be differentiable so that gradient-based methods can still be applied, and that such gradients can be efficiently computed.

    % Generally nonlinear problems of the form in \ref{pblm:ct} must be solved numerically, and we replace the continuous-time objective with a discretized form
    % \begin{equation}\label{pblm:dt}
    %   \min \quad g_T(x_K, u_K) + \sum_{k=1}^K g(x_k, u_k) + \tr \big( \mx{Y} \mx{P}_k \mx{Y}^\T \big)
    % \end{equation}
    % representing approximate integration over $K$ sequential time intervals $[t_k, t_{k+1}]$.
    % The discrete-time covariance update equations (corresponding to the optimal estimator over the LTV system (\ref{eq:dyn_ltv})) can be written
    % where (\ref{eq:ekf_2}) shows that the Fisher information (\ref{eq:fisher_info}) is, in fact, captured by the posterior covariance.
    % As will be assumed in Section \ref{s:bundling}, many real-world sensors render measurements
    % at discrete time intervals, in which case a discrete-time EKF is appropriate and not simply a numerical convenience.

  \subsection{Choices of Uncertainty Metric $\Jobs$}
    At a high level, a good choice of uncertainty term $\Jobs$ should produce a multi-objective optimization which smoothly and efficiently trades expedience (captured by $J_c$) for estimation performance.
    A number of choices for $\Jobs$ have been proposed in the literature, but they generally fall into one of three classes.

    %\begin{itemize}
      \subsubsection{Maximizing landmark visibility:}
      %\item \textbf{Maximizing landmark visibility:}
      A common heuristic in the case of landmark-based estimators (i.e.\ visual-inertial odometry) is to maximize some visibility metric~\cite{falanga2018pampc,murali2019perception}.
      While this encourages onboard sensors to be pointed towards informative parts of the environment, it does not explicitly capture the observability properties of the system.

      \subsubsection{Maximizing the Observability Gramian or Fisher-information:}
      %\item \textbf{Maximizing the Observability Gramian or Fisher-information:}
      From a control-theoretic perspective,~\cite{krener2009measures} and~\cite{preiss2018simultaneous} propose maximization of metrics based on the Local Observability Gramian (LOG).
      The LOG is equivalent to the Fisher information up to a constant scaling.
      Because $\Lam \succeq 0$ is singular for trajectories about which the system is locally unobservable, \cite{krener2009measures} and \cite{preiss2018simultaneous} propose maximization of the smallest eigenvalue $s_1(\Lam) \geq 0$.
      %In order to capture the nonlinear dynamics and improve the rank of $\Lam$ to better approximate the true observability, \cite{preiss2018simultaneous} propose a Taylor-series approximation based on the Lie derivatives of the measurement model $h$.
      %Our bundling approach, described in Section \ref{s:bundling}, is similar in spirit although explicitly formulated for a discrete-time sensor and interpreted differently.

      However, there are some challenges to direct maximization.
      Each row and column of $\Lam$ corresponds to a different estimated state variable, and can refer to quantities as varied as positions, velocities, and IMU biases.
      Maximization of individual sub-matrices (e.g., the position block) as proposed by \cite{preiss2018simultaneous} maintains consistent units of measurement, but is information-theoretically equivalent to \emph{conditioning} on all other states (treating them all as known) and fundamentally neglects key observability properties of the system.
      Joint maximization, on the other hand, requires some scaling method; two different statistical approaches are presented in \cite{preiss2018simultaneous} and \cite{nobre2019learning}.
      Ultimately, direct maximization of $\Lam$ is heuristic and, as pointed out by \cite{rafieisakhaei2017use}, can produce expensive trajectories (with respect to $J_c$) that yield little improvement in actual estimation error.

      \subsubsection{Minimizing posterior covariance:}
      %\item \textbf{Minimizing posterior covariance:}
      Following a number of existing works~\cite{prentice2009belief,van2011lqg}, we propose minimization of the posterior estimator covariance $\mx{P}(t) \succeq 0$.
      In contrast to the Fisher information or LOG, the posterior covariance of the EKF captures uncertainty in both the system dynamics and sensing.
      Furthermore, sub-blocks of $\P_k$ represent the \emph{marginal} covariances over those variables, capturing the effects of all other unknown states.
      Minimization over the trajectory allows the optimization to smoothly trade-off between minimizing uncertainty and conventional planning costs, even to the extent of allowing $\Lam_k$ singular at some instances.
      Moreover, this trade-off can naturally take into account the initial uncertainty $\P_0$.

      In particular, if $\mx{Y}$ is chosen to select the sub-block of $\P_k$ corresponding to the estimated position $\vc[w]{\hat{p}}_k$, then
      \begin{equation}\label{eq:junc_pc}
        %\mx{Y} = \begin{bmatrix} \mx{I}_d & \mx{0} \end{bmatrix} \, \implies \, \Jobs(u) = \mathsmaller{\sum_{k=1}^K} \tr \big( \mx{Y} \mx{P}_k \mx{Y}^\T \big) = \mathsmaller{\sum_{k=1}^K} \E ||\hat{\delta p}_k - \delta p_k||_2^2
        %\mx{Y} = \begin{bmatrix} \mx{I}_d & \mx{0} \end{bmatrix} \, \implies \, \Jobs(u) \triangleq \sum_{k=1}^K \tr \big( \mx{Y} \mx{P}_k \mx{Y}^\T \big) = \sum_{k=1}^K \E ||\hat{\delta p}_k - \delta p_k||_2^2
        \Jobs(u) \triangleq \sum_{\kbar=1}^{\Kbar} \tr \big( \mx{Y} \mx{P}_\kbar \mx{Y}^\T \big) = \sum_{\kbar=1}^\Kbar \E ||\vc[w]{\hat{p}}_\kbar - \vc[w]{p}_\kbar||_2^2.
      \end{equation}
      That is, this choice of $\Jobs$ explicitly minimizes the mean-squared estimator error over position (with well-defined units of m$^2$).
      In a sense, minimizing the posterior covariance is similar to maximizing $\Lam$, but warped and scaled \emph{correctly} by the dynamics and prior uncertainty of the system.
  %\end{itemize}

%  \subsection{Existing challenges}
%    In order to optimize (\ref{eq:junc_pc}) in an online motion-planning context, we must be able to evaluate it (and compute gradients) efficiently.
%    However, the covariance update equations (\ref{eq:ekf_1}) and (\ref{eq:ekf_2}) (and equivalent forms) have at least quadratic complexity in the dimension of $\P$.
%    Thus, simulating the EKF exactly over a given trajectory can be expensive when the sensor rate is high (requiring many updates) or in the presence of a large number of unknown landmarks (requiring large $\P$).
%
%    We address these challenges by 1) bundling high-rate measurements together, 2) marginalizing out unknown landmark parameters, and 3) collapsing measurement contributions from multiple landmarks into a compact quadratic form.
%    These modest approximations are shown to dramatically reduce the computational cost of evaluating (\ref{eq:junc_pc}).

\section{Approximating the SLAM Estimation Process}\label{s:interval_filter}
  In order to solve the optimal control problem over the continuum of possible control inputs, we require an efficient means of computing posterior covariance as a function of a nominal trajectory.
  This section outlines our approximation and how it differs from existing work, and introduces the proposed Structureless Interval Information Filter (SIIF) framework that will be elaborated in the rest of the paper.

  A key aim of this paper is to efficiently approximate the \emph{covariance dynamics} corresponding to the locally-LTV system (\ref{eq:ltv}), in order to enable observability-aware trajectory generation.
  Note that, if it were not for the dependence on the environmental features $\{\elln\}$, these dynamics would be given by the familiar Kalman Filter (KF) equations.
  While the point (mean) state estimate evolves according to the realized observations (and is therefore unknown at planning time), the error covariance evolves \emph{deterministically} as a function of the underlying, linearizing trajectory
  \footnote{We distinguish the (linear) KF defined according to the given LTV dynamics from the Extended KF (EKF) more often used in practice.
            Unlike the KF, the EKF \emph{re}-linearizes the nonlinear dynamics based on its current state estimate, and thus the ensuing Jacobians and posterior covariance are indeed dependent on the random observations.
            In contrast, the determinism of the KF covariance makes it much more useful for the purposes of observability-aware planning.}.

  \subsection{Challenges in the SLAM case}
    \begin{figure}[h]
      \centering
      \subfloat[Typical SLAM factor graph\label{fig:slam_graph}]{
        \includegraphics[width=\textwidth]{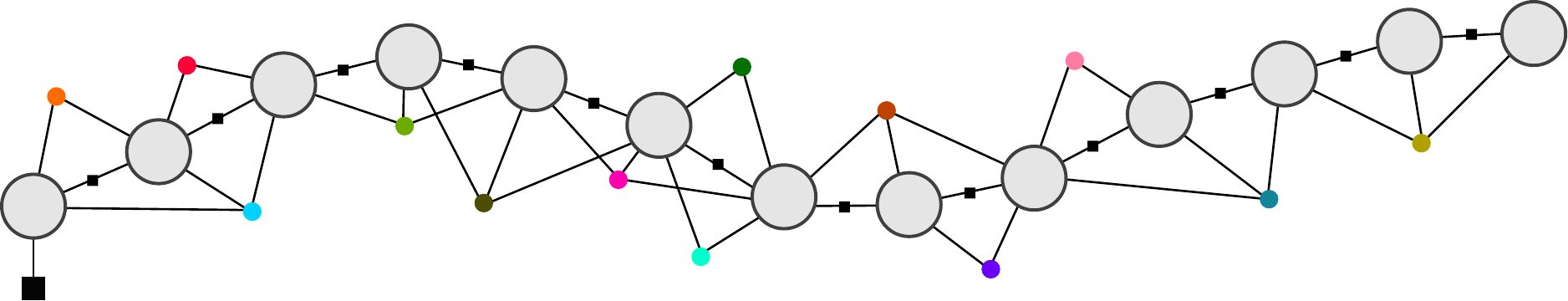}
      }\newline
      \subfloat[Our Structureless Interval Filter (SIIF) approximation\label{fig:interval_filter}]{
        \includegraphics[width=\textwidth]{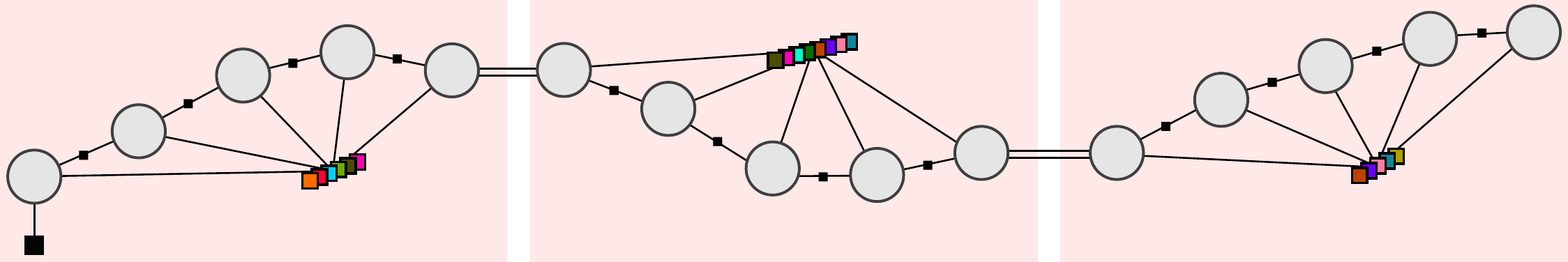}
      }
      \caption{The original SLAM factor graph includes both pose and landmark variables over the trajectory, and its structure depends on FoV constraints and (potentially unknown) landmark locations.
               Even when the graph is fully known at planning time, exact computation of the posterior covariance is expensive, prohibiting the multiple evaluations required for online observability-aware planning.
               In contrast, our SIIF approximation breaks up the trajectory into time intervals and discards inter-interval loop closures,
               Then connectivity can be generalized by enforcing a landmark visibility weighting within the Jacobians themselves, and landmarks can then be marginalized out over each interval.
               This produces a compressed sequence of \emph{structureless}, ``bundled'' measurements allowing the covariance to be propagated efficiently according to a recursive update law, as described in Section \ref{s:siif_update}.
               }%Thus our method avoids the limitations imposed by reasoning over factor graphs or maintaining the information state of the uncertain landmarks.}
    \end{figure}

    When sensors \emph{do} depend on landmarks as in (\ref{eq:obs}) -- the setting assumed in this paper -- predicting and optimizing estimator performance over trajectories becomes fundamentally more challenging.
    This is because the locations of landmarks are usually uncertain (the case assumed in SLAM), and indeed often completely unknown at planning time.
    Of course, a number of filtering approaches such as MonoSLAM \cite{davison2007monoslam} and ROVIO \cite{bloesch2017iterated} account for landmark uncertainty by including the landmark error $\{\dlmn\}$ in the error state.
    However, this augmentation comes at a quadratic cost with the number of landmarks and is clearly intractable when numerous rapid evaluations are required for online optimization.

    A popular alternative to filtering approaches instead views the estimation problem over the entire trajectory (\ie, in batch) as a factor graph, as in Fig.~\ref{fig:slam_graph}.
    In a factor graph, \emph{variable} nodes represent unknowns such as robot poses and landmark locations, and \emph{factor} nodes represent measurements or priors.
    Significant work over the last decade has produced efficient estimation techniques over factor graphs, particularly leveraging sparsity \cite{dellaert2006square,kaess2011isam2,elimelech2017scalable,kopitkov2017no} to minimize computation.
    However, these techniques assume full knowledge of the graph, which ultimately depends on sensor field-of-view (FoV) and therefore on both the trajectory and locations of landmarks.
    When landmark locations are unknown, observability-aware trajectory optimization could then be viewed as an optimization over (random) graphs -- a clearly unsuitable formulation in the context of online, continuous trajectory optimization
    \footnote{When landmark locations \emph{are} approximately known, approaches like \cite{elimelech2017scalable,kopitkov2017no,zhang2018perception} nevertheless do adopt the search-over-graphs formulation.
              However, their methods are restricted to evaluation of a finite set of candidate trajectories.}.

  \subsection{The SIIF Formulation}
    To avoid the complexities of the graph-theoretic view, we adopt a ``structureless'' filtering formulation similar to the MSCKF \cite{mourikis2007multi}.
    In order to properly account for landmark uncertainty across timesteps \emph{without} augmenting the error state vector, we develop a bundled filter update over time \emph{intervals}.
    As illustrated in Fig.~\ref{fig:interval_filter}, this can be visualized by explicitly partitioning the factor graph according to these \emph{a priori} time intervals and then marginalizing the landmarks.
    The resulting estimation problem can be solved algorithmically via a recursive, bundled updating procedure we refer to as the \emph{interval filter}.

    It is crucial to note here that for the purpose of optimizing observability-aware trajectories, we are only concerned with efficiently \emph{representing} the estimation process, and are not actually producing an estimate.
    To accomplish this we only require a (deterministic) model of covariance propagation over time and do not need to produce an actual state estimate (as this would require access to ``future'' observations).
    Moreover, the interval filtering model proposed in Fig.~\ref{fig:interval_filter} is clearly sub-optimal, as it decouples inter-interval connections and loop closures (thus discarding information).
    Nevertheless, it will be seen that the computational and algorithmic benefits of this approximation are significant, and the resulting model is sufficient to capture the ``short-horizon'' observability characteristics of SLAM for the purposes of planning.
    Furthermore, this will enable generalization to the case where landmark locations are only known in a distributional sense, and thus estimator performance can be predicted well beyond known space.

    %This explicit partitioning of the graph into intervals discards loop closures, and therefore induces some information loss.
    %However, this approximation is conservative in an information-theoretic sense, such that the resulting uncertainty estimate is guaranteed to be an \emph{over}-estimate.
    %Furthermore, it maintains the fundamental observability properties of SLAM over short horizons as an ``odometry sensor,'' and for receding-horizon trajectory optimization (which is unlikely to be concerned with loopy trajectories), this loss is minimal.

    The remainder of this paper details how this interval filter is propagated, how landmarks can be anticipated into the future, and how the resulting covariance estimate can be used in planning.
    Section \ref{s:siif_update} describes how measurements over each interval can be bundled and the landmarks marginalized out.
    Additionally, it proposes a Lie-derivative-based approximation which greatly simplifies Jacobian formation in the case of high-rate sensors such as cameras.
    Section \ref{s:affine_models} defines a privileged class of sensor models (those with Jacobians affine in the landmark $\lmn$), and describes how information contributions from all landmarks can be aggregated directly without explicit (linear-time) summation.
    Besides expediting computation, this allows for generalization to \emph{distributions} of landmarks, allowing natural reasoning about where new landmarks might be discovered.
    Finally, Section \ref{s:results} gives some preliminary planning results demonstrating that significant covariance reduction can indeed be achieved for popular systems such as quadrotor-equipped VIO, and that heuristic alternatives are often ineffective.

\section{The SIIF Update}\label{s:siif_update}
  For high-rate sensors such as cameras, simulation of the EKF update (and subsequent gradient back-propagation) at measurement rate can be intractable for real-time planning.
  For this reason, we propose a computationally-efficient method to ``bundle'' multiple high-rate measurements into a single update.
  In contrast to the traditional Kalman Filter, which proceeds one measurement timestep at a time, our approximate filter operates on larger ``chunks'' of time, incorporating batches of observations according to user-specified interval length $K > 1$.
  The overall planning horizon is comprised of $\Kbar$ such intervals, accounting for a total of $K \Kbar$ individual measurement timesteps.
  Besides reducing the total number of ``updates'' required to model high-rate sensors, this allows for landmark uncertainty to be marginalized away within each interval, avoiding the need to track landmark uncertainty \emph{across} intervals.
  This marginalization is information-theoretically ``lossy'' but yields significant computational and algorithmic benefits while capturing the fundamental observability characteristics of the system, particularly over short horizons.
  %This provides a simplified proxy for the true onboard estimation scheme, facilitating efficient evaluation and optimization of trajectories and allowing for landmark marginalization across short time horizons.

  Without loss of generality, assume that measurements are collected every $\Dt$ seconds and consider a single time interval $[0, T = K \Dt]$ for some $K > 1$.
  Let $\pT = \{ t_0 = 0, t_1, t_2, \ldots, t_k = T \}$ refer to the set of corresponding $(K+1)$ evenly-spaced timesteps.
  Following a conventional filtering paradigm, assume we are given a prior estimate $\xhat_0$ with corresponding error $\e_0 \sim \MVN(\vc{0}, \P_0 = \S_0^{-1})$.
  From here, we seek an update equation or algorithm that ingests the proceeding \emph{batch} of linearized observations $\big( \zntildek[1], \zntildek[2], \ldots, \zntildek[K] \big)$ and outputs a posterior estimate $\xhat_K$.
  However, it is worth pointing out that for the purposes of planning, we do not actually need to do inference (that is, compute the point estimate $\xhat_K$).
  Rather, in this paper we focus solely on the covariance update, which as discussed in the previous section can indeed be considered a deterministic function of the nominal trajectory.

  From the continuous-time LTV dynamics (\ref{eq:ltv}), we can produce a discrete-time error dynamics model
  \begin{align}
    \e_{k+1}   &= \underbrace{ \big(\mx{I} + \Dt \mx{A}_c(x_k, u_k) \big) }_{\triangleq \mx{A}_k } \e_k + \underbrace{ \sqrt{ \Dt } \mx{G}_c(x_k, u_k) }_{\triangleq \mx{G}_k} \w_k   \label{eq:ltv_error} \\
    \zntildek  &= \sigma(x_k, \elln) \underbrace{ \mx{H}(x_k; \elln) }_{\triangleq \mx{H}_k^{(n)}} \e_k + \underbrace{ \mx{L}(x_k; \elln) }_{\triangleq \mx{L}_k^{(n)}} \lmntilde + \nunk  \label{eq:ltv_obs}
  \end{align}
  where $\sigma(x; \elln) \in [0,1]$ indicates the ``visibility'' of landmark $\elln$ from state $x$.
  When $\sigma(x; \elln) = 0$ (\ie, the landmark is out of view), and no information is obtained.
  Under evaluation, $\sigma$ takes ``hard'' values of $\{0,1\}$, but during optimization it will be useful to instead apply a smooth approximation as in \cite{zhang2019beyond}.
  A candidate choice of $\sigma$ tailored for use with pinhole camera models is given in Appendix \ref{app:fov}.

  For the purposes of our interval filtering approximation, it will be useful to assume that visibility is \emph{constant} over each interval -- that is, we will replace (\ref{eq:ltv_obs}) with
  \begin{equation}\label{eq:ltv_obs_constant_viz}
    \zntildek  \approx \sigma(x_0, \elln) \underbrace{ \mx{H}(x_k; \elln) }_{\triangleq \mx{H}_k^{(n)}} \e_k + \underbrace{ \mx{L}(x_k; \elln) }_{\triangleq \mx{L}_k^{(n)}} \lmntilde + \nunk.
  \end{equation}
  Though clearly approximate, this will allow significant computational speedup and is intuitively reasonable for short interval lengths.

  \subsection{Overview: Update in information space}
    From (\ref{eq:ltv_error}) it is straightforward to write the error state \emph{trajectory} over the interval
    \begin{equation} \label{eq:ltv_batch_phi}
      \begin{bmatrix} \e_1 \\ \e_2 \\ \vdots \\ \e_K \end{bmatrix}
        =
      \underbrace{ \begin{bmatrix} \mx{A}_0 \\ \mx{\Phi}^1_0 \\ \vdots \\ \mx{\Phi}^{K-1}_0 \end{bmatrix} }_{\triangleq \Ablock(x_0, u_0)}
      \e_0
        +
      \underbrace{ \begin{bmatrix} \mx{G}_0 & & & & \\  \mx{A}_1 \mx{G}_0 & \mx{G}_1 & & & & \\  \mx{\Phi}^2_1 \mx{G}_0 & \mx{A}_2 \mx{G}_1 & \mx{G}_2 & & & \\  \vdots & \vdots & \vdots & \ddots & \end{bmatrix} }_{\triangleq \Gblock(x_0, u_0)}
      \underbrace{ \begin{bmatrix} \vc{w}_0 \\ \vc{w}_1 \\ \vc{w}_2 \\ \vdots \\ \vc{w}_{K-1} \end{bmatrix} }_{\triangleq \W}
    \end{equation}
    where the error state transition matrix $\mx{\Phi}^{k_1}_{k_2} \triangleq \mx{\Phi}(t_{k_1}, t_{k_2}) \approx \mx{A}_{k_1} \mx{A}_{k_1-1} \ldots \mx{A}_{k_2}$.

    Using (\ref{eq:ltv_obs_constant_viz}), the corresponding measurement innovations for each landmark $\elln$ can be written
    \begin{align}
      \begin{bmatrix} \zntildek[1] \\ \zntildek[2] \\ \vdots \\ \zntildek[K] \end{bmatrix}
        &=
        \sigma(x_0; \elln) \Bigg(
        \underbrace{ \blkdiag \begin{bmatrix} \mx{H}_1^{(i)} \\ \mx{H}_2^{(i)} \\ \vdots \\ \mx{H}_K^{(i)} \end{bmatrix} }_{\triangleq \Hblock(x_0, u_0; \elln)}
        \begin{bmatrix} \e_1 \\ \e_2 \\ \vdots \\ \e_K \end{bmatrix}
          +
        \underbrace{ \begin{bmatrix} \mx{L}_1^{(i)} \\ \mx{L}_2^{(i)} \\ \vdots \\ \mx{L}_K^{(i)} \end{bmatrix} }_{\triangleq \Lblock(x_0, u_0; \elln)}
        \lmntilde
      \Bigg)
        +
      \underbrace{ \begin{bmatrix} \nunk[1] \\ \nunk[2] \\ \vdots \\ \nunk[K] \end{bmatrix} }_{\triangleq \Nun}
      \label{eq:constant_viz} \\
        &=
      \sigma_0^{(n)} \Big( \Hnblock \begin{bmatrix} \Ablock & \Gblock \end{bmatrix} \begin{bmatrix} \e_0 \\ \W \end{bmatrix}
                              + \Lnblock \lmntilde \Big)
                      + \Nun  \label{eq:ltv_batch_H}
    \end{align}
    where for clarity we've introduced the shorthand $\Hnblock = \Hblock(x_0, u_0; \elln)$ and $\Lnblock(x_0, u_0; \elln)$ and otherwise suppressed the $(x_0, u_0)$ dependence.

    As shown by Mourikis \etal \cite{mourikis2007multi}, the landmark error $\lmntilde$ can be marginalized out by computing the left-nullspace matrix $\Qn$ subject to
    \begin{equation}\label{eq:null_Qn}
      \Qn^\T \Lnblock = \mx{0} \quad \text{ and } \quad \Qn^\T \Qn = \mx{I}.
    \end{equation}
    Then left-multiplication of (\ref{eq:ltv_batch_H}) by $\Qn^\T$ produces a \emph{marginalized} innovation
    \begin{equation}\label{eq:explicit_innovation}
      \Znexp \triangleq \sigma_0^{(n)}
              \underbrace{ \Qn^\T \Hnblock \begin{bmatrix} \Ablock & \Gblock \end{bmatrix} }_{\triangleq \Hbarblockexp(x_0, u_0; \elln)}
              \begin{bmatrix} \e_0 \\ \W \end{bmatrix}
            + \Qn^\T \Nun
    \end{equation}
    (\ref{eq:explicit_innovation}) allows us to interpret the sequence of measurements corresponding to $\elln$ as a large, joint measurement over $\e_0$ and $\W$.
    Note that by construction, the $\Qn^\T \Nun$ term refers to an i.i.d. additive noise with unit covariance.
    This perspective allows the joint information over $(\e_0, \W)$ to be combined additively as
    \begin{equation}\label{eq:explicit_lambda_sum}
      \Lam_0(x_0, u_0) =
      \underbrace{ \begin{bmatrix} \S_0 & \mx{0} \\  \mx{0} & \mx{I} \end{bmatrix} }_{\triangleq \Lam_0^{\text{prior}}}
        +
      \mathlarger \sum_{n = 1}^{N} \sigma_0^{(n),2} \, \underbrace{ \Hbarblockexp(x_0, u_0; \elln)^\T \Hbarblockexp(x_0, u_0; \elln) }_{\triangleq \Lam_0^{\mathcal{Z}}(x_0, u_0; \elln)}.
    \end{equation}
    In (\ref{eq:explicit_lambda_sum}) we recall the functional dependence of the underlying Jacobians and the visibility score $\sigma$ on the nominal trajectory over the interval, which in turn is a deterministic function of $x_0$ and the control over the interval $u_0$.
    Furthermore, note that the information matrices involved are defined jointly over a \emph{single} error state $\e_0$ and the interval-length process noise $\W$.

    While (\ref{eq:explicit_lambda_sum}) gives a convenient form for the information ``update'', we still must propagate the uncertainty from initial state $\e_0$ to the error state at the end of the interval $\e_K$ and marginalize out the nuisance variable $\W$.
    Fortunately, we have from (\ref{eq:ltv_error}) we have
    \begin{equation*}
      \begin{bmatrix} \e_K \\ \W \end{bmatrix}
        =
      \underbrace{
      \begin{bmatrix} \mx{\Phi}^K_0 & \mx{\Phi}^K_1 \mx{G}_0 & \mx{\Phi}^K_2 \mx{G}_1 & \ldots & \mx{G}_{K-1}  \\
                      \mx{0}        &  &  \mx{I} & &
      \end{bmatrix}
      }_{\triangleq \blockM(x_0, u_0)}
      \begin{bmatrix} \e_0 \\ \W \end{bmatrix}
    \end{equation*}
    and it is then straightforward to verify that
    \begin{equation}\label{eq:info_propogate}
      \Lam_K = \blockM^{-\T} \Lam_0 \blockM^{-1}.
    \end{equation}
    Note that inversion of $\blockM$ is possible so long as $\mx{\Phi}^K_0$ is invertible (a very mild condition), and can be significantly accelerated by leveraging the block sparsity pattern.
    From $\Lam_K$, the marginal covariance over $\e_k$ can be extracted via Schur complement
    \begin{equation}\label{eq:W_marginalize}
      \S_K = [ \Lam_K ]_{\e_K,\e_K} - [ \Lam_K ]_{\e_K, \W} [ \Lam_K ]^{-1}_{\W, \W} [ \Lam_K ]_{\W, \e_K}.
    \end{equation}

    Taken together, (\ref{eq:ltv_batch_phi}-\ref{eq:W_marginalize}) describe an exact (up to the constant visibility assumption) information update under our ``interval filter'' paradigm.
    Crucially, we avoid ever explicitly representing the uncertainty over landmarks, and as long as the visibility function $\sigma(x_0, u_0)$ is a smooth function of the initial state and interval control, $\S_K = \S_K(x_0, u_0)$ will be also.

    However, there are a few limitations of this ``explicit'' approach.
    First, dimensionality of the batch Jacobians in (\ref{eq:ltv_batch_phi}) and (\ref{eq:ltv_batch_H}), and thus the dimensionality of $\W$ and $\Lam_0$, grows with the size of the interval, and this multi-step integration represents non-trivial computation.
    Secondly, and more relevant to the SLAM problem, evaluation of $\Lam_0$ according to (\ref{eq:explicit_lambda_sum}) requires explicit formation and summation of information contributions from each of the $N$ landmarks.
    For observability-aware motion planning, this additionally requires knowledge of the linearization points $\{\elln\}$ at planning time, a restrictive assumption that prohibits anticipation into unknown space.
    The remainder of this section will address the first issue by introducing a Lie-derivative-based approximation of $\LamZ_0(x_0, u_0; \elln)$, and a generalization beyond the finite landmark set $\{\elln\}$ will be the subject of Section \ref{s:affine_models}.

    \subsection{A Lie-Taylor approximation of $\LamZ_0$}\label{s:lie_taylor_update}
      In this section we introduce a Taylor-style approximation of $\LamZ_0$ based on the Lie derivatives of $h\big( x(t); \elln \big)$.
      This approach effectively approximates the batch Jacobians (\ref{eq:ltv_batch_phi}) and (\ref{eq:ltv_batch_H}) in a fixed-dimensional representation (\ie, independent of interval length $K$).
      In doing so, the interval process noise $\W$ is approximated in a fixed-dimensional form -- this is naturally desirable, as $\W$ is ultimately a ``nuisance'' variable modeling an abstract noise source, and ``exact'' representation is neither well-defined nor necessary.
      Ultimately, our approach here can be considered a stochastic extension of the approximation proposed in a slightly different context by Preiss \etal \cite{preiss2018simultaneous}.

      Let $\Ljh(x_0, u_0; \elln)$ represent the $j$-th Lie derivative of $h\big(x(t); \elln \big)$ with respect to the nominal dynamics $f$ taken at $x(0) = x_0$.
      Then for small $t > 0$, we can approximate the instantaneous measurement model as
      \begin{equation}\label{eq:lie_taylor_obs}
        h\big(x(t); \elln\big) \approx \sum_{j = 0}^{r-1} \frac{t^j}{j!} \Ljh(x_0, u_0; \elln)
      \end{equation}
      using the first $r$ Lie derivatives taken at $x_0$.

      However, the Lie derivatives $\{\Ljh\}$ take into account only the nominal dynamics and neglect any form of process uncertainty.
      To address this, we would like to find a finite-dimensional representation of Brownian noise process $\w(t)$ as it appears in the continuous diffusion (\ref{eq:dyn}).
      However, the continuous-time process $\w(t)$ is of course infinite-dimensional.
      Instead, we locally model the continuous-time error dynamics via a particular jump process with a ``matching'' discrete-time approximation.
      This effectively amounts to applying a fixed-dimensional process noise ``all-at-once'' as an instantaneous ``jump'' at $t_0$.
      More specifically, say that at time $t_0 = 0$,
      \begin{equation}\label{eq:perturbed_e_def}
        \e_0^+ = \e_0 + \sqrt{t_K} \mx{\Phi}_K^0(x_0, u_0) \mx{G}_c(x_0, u_0) \w_0
                = \underbrace{ \begin{bmatrix} \mx{I} \,&\, \sqrt{t_K} \mx{\Phi}_K^0 \mx{G}_c \end{bmatrix} }_{\triangleq \mx{E}(x_0, u_0)} \begin{bmatrix} \e_0 \\ \w_0 \end{bmatrix}
      \end{equation}
      where $\w_0 \sim \MVN(\vc{0}, \mx{I})$ as before.
      From our discrete-time dynamics (\ref{eq:ltv_error}) we can verify
      \begin{equation}
        \e_K \approx \mx{\Phi}_0^K \e_0 + \sqrt{t_K} \mx{G}_c \w_0 = \mx{\Phi}_0^K \e_0^+,
      \end{equation}
      suggesting that (\ref{eq:perturbed_e_def}) is a reasonable perturbation model.
      Of course, the stochastic jump at $t_0$ is zero-mean, and therefore the linearization point $x_0^+ = x_0$ is unaffected.

      Now, from (\ref{eq:lie_taylor_obs}) we can write the instantaneous measurement Jacobian with respect to the \emph{perturbed} initial error state $\e_0^+$ as a sum of the Jacobians of the corresponding Lie derivatives
      \begin{align}
        \zntildek
          &=
            \sigma_0^{(n)} \sum_{j = 0}^{r-1} \frac{t_k^j}{j!}
            \begin{bmatrix} \Hljn & \Lljn \end{bmatrix}
            \begin{bmatrix} \e_0^+ \\ \lmntilde \end{bmatrix}
              + \nunk  \nonumber \\
          &=
            \sigma_0^{(n)} \sum_{j = 0}^{r-1} \frac{t_k^j}{j!}
            \begin{bmatrix} \Hljn \mx{E} \,&\, \Lljn \end{bmatrix}
            \begin{bmatrix} \e_0 \\ \w_0 \\ \lmntilde \end{bmatrix}
              + \nunk   \label{eq:lie_nablas}
      \end{align}
      where $\Hljn = \Hlj(x_0, u_0; \elln)$ and $\Lljn = \Llj(x_0, u_0; \elln)$ are the Jacobians of the $i$-th Lie derivative with respect to $\e_0^+$, $\lmntilde$, respectively.
      Note that use of Lie derivatives as in (\ref{eq:lie_nablas}) is particularly attractive because it avoids the need for explicit (and linearized) integration Jacobians as in (\ref{eq:ltv_batch_phi}).

      As before, we need to combine the measurements $\{\zntildek\}$ over the all $k$ and marginalize out the landmark error $\lmntilde$.
      Rather than doing this explicitly, first note that the total information over $(\e_0^+, \lmntilde)$ can be expressed
      \begin{align}
        &\, \sum_{k = 1}^K \sum_{i = 0}^{r-1} \sum_{j = 0}^{r-1} \Big( \frac{ t_k^i }{i!} \begin{bmatrix} \Hljn[i] & \Lljn[i] \end{bmatrix} \Big)^\T  \Big( \frac{ t_k^j }{j!} \begin{bmatrix} \Hljn & \Lljn \end{bmatrix} \Big)  \nonumber \\
        &= \begin{bmatrix} \Hljn[0] & \Lljn[0] \\ \Hljn[1] & \Lljn[1] \\ \vdots & \vdots \\ \Hljn[r-1] & \Lljn[r-1]  \end{bmatrix}^\T
           %\underbrace{ \begin{bmatrix}
           %         \lambda_0 & \lambda_1 & \frac{\lambda_2}{2} & \ldots & \frac{\lambda_{r-1}}{(r-1)!}  \\
           %         \lambda_1 & \lambda_2 & \frac{\lambda_3}{2} & \ldots & \frac{\lambda_r}{(r-1)!}  \\
           %         \frac{\lambda_2}{2} & \frac{\lambda_3}{2} & \ddots & & \frac{\lambda_{2(r-1)}}{2(r-1)!}  \\
           %         \vdots  \\
           %         \frac{\lambda_{r-1}}{(r-1)!} & & & & \frac{\lambda_{2r-1}}{(r-1)!^2}
           %             \end{bmatrix}
           % }_{\triangleq \mx{W}(T)}
           \big( \mx{W} \kron \mx{I}_{m \times m} \big)
           \begin{bmatrix} \Hljn[0] & \Lljn[0] \\ \Hljn[1] & \Lljn[1] \\ \vdots & \vdots \\ \Hljn[r-1] & \Lljn[r-1]  \end{bmatrix}  \label{eq:info_lie_stacking}
      \end{align}
      where $\kron$ represents the Kronecker product.
      The $r \times r$ coupling matrix $\mx{W}$ is defined element-wise as
      \begin{equation}
        \mx{W}(\pT)_{ij} = \frac{\lambda_{i+j}(\pT)}{i!j!}  \quad \text{ and } \quad \lambda_s(\pT) \triangleq \sum_{k = 1}^K (t_k)^s
      \end{equation}
      where indices $i,j$ run from 0 to $(r-1)$ (\ie, they are zero-indexed).
      The form (\ref{eq:info_lie_stacking}) indicates that by expressing the measurement model as a finite sum of its Lie derivatives, the combination of measurements over the \emph{interval} can be expressed as a \emph{stacked} Jacobian of finite dimension.
      The magnitude and couplings between contributions from different derivatives is determined by the measurement timestamps $\pT$ and captured by $\mx{W}(\pT)$, whereas the rank and observability structure of the information contribution are determined by $\Hljn$ and $\Lljn$.
      Crucially, the inclusion of higher-order derivatives generally increases the rank of the bundled measurement, capturing the fact that states that are unobservable under a single measurement may become observable over multiple sequential sensor readings.

      Letting the marginalization matrix $\Qn$ be defined according to
      \begin{equation}
        \Qn^\T (\mx{W} \kron \mx{I})^{\frac{1}{2}} \begin{bmatrix} \Lljn[0] \\ \Lljn[1] \\ \vdots \\ \Lljn[r-1] \end{bmatrix} = \mx{0}
          \quad \text{ and } \quad \Qn^\T \Qn = \mx{I}
      \end{equation}
      we can define a Lie-Taylor based measurement residual as
      \begin{equation}\label{eq:lie_innovation}
        \Znlie = \underbrace{ \Qn^\T (\mx{W} \kron \mx{I})^{\frac{1}{2}} \begin{bmatrix} \Hljn[0] \\ \Hljn[1] \\ \vdots \\ \Hljn[r-1] \end{bmatrix} \mx{E} }_{\triangleq \Hbarblocklie(x_0, u_0; \elln)}
              \begin{bmatrix} \e_0 \\ \w_0 \end{bmatrix}
              +
              \Qn^\T \Nun.
      \end{equation}
      Note that $\mx{W}(\pT)$ is independent of $x_0$ and $\elln$, and thus the matrix square root is constant and can be pre-computed.

      From (\ref{eq:lie_innovation}) the information contribution $\LamZ_0(x_0, u_0, \elln)$ over $(\e_0, \W = \w_0)$ can be computed for each $\elln$, and the update and propagation can proceed as before.
      The overall update-propagation procedure for the interval filter is given in Algorithm \ref{algo:interval_update}.

      %  In contrast to the E$^2$LOG presented by \cite{preiss2018simultaneous}, the bundled measurement (\ref{eq:taylor_lambda}) explicitly approximates a high-rate \emph{discrete}-time sensor (like a camera).
      %  Furthermore, unlike \cite{preiss2018simultaneous} we do not seek to maximize this quantity directly.

      \begin{algorithm}
        \SetAlgoLined
        \DontPrintSemicolon
        \KwInput{Nominal state and control $x_0, u_0$}
        \KwInput{$N$ landmark linearization points $\{ \elln \}$}
        \KwInput{Measurement timestamps $\pT = \{t_0, t_1, \ldots, t_K \}$}
        \KwInput{Prior info matrix $\S_0 = \P_0^{-1}$ over $\e_0$}
        \KwOutput{Posterior info matrix $\mx{S}_K = \P_K^{-1}$ over $\e_K$}

        \tcc{Build $\Lam_0$}
        $\Lam_0 \gets \begin{bmatrix} \mx{S}_0 & \mx{0} \\ \mx{0} & \mx{I} \end{bmatrix}$
        \tcp*{init joint info over $(\e_0, \W)$}

        \tcc{For each landmark}
        \For{$n$ in $\{1, 2, \ldots, N\}$} {

          $\sigma_0 \gets \sigma(x_0, \elln)$
          \tcp*{compute visibility}

          %\tcc{If visible from $x_0$}
          \If{$\sigma_0 > 0$} {
            \tcc{Info contribution via explicit or Lie-Taylor method.}
            $\LamZ \gets \textsc{ComputeInfoContribution}(\pT, x_0, u_0, \elln)$

            $\Lam_0 \gets \Lam_0 + \sigma_0^2 \LamZ$
          }
        }

        \tcc{Propagate and marginalize}
        $\Lam_K \gets \blockM^{-\T} \Lam_0 \blockM^{-1}$
        \tcp*{see (\ref{eq:info_propogate}), can be implemented sparsely}
        %$\mx{S}_K \gets [\Lam_K]_{\e,\e} - [\Lam_K]_{\e,\W}[\Lam_K]_{\W,\W}^{-1} [\Lam_K]_{\W,\e}$
        $\mx{S}_K \gets \textsc{SchurComplement}(\Lam_K)$
        \tcp*{see (\ref{eq:W_marginalize})}

        \caption{Interval Filter Update}
        \label{algo:interval_update}
      \end{algorithm}

\section{Handling Many Landmarks}\label{s:affine_models}
  The preceding discussion assumed a generic (albeit sufficiently differentiable) per-landmark observation function $h(x; \ell)$.
  Moreover, it is clear from Algorithm \ref{algo:interval_update} that the measurement bundling and landmark marginalization approximations as stated will still assume an explicit (and finite) set of landmark linearization points $\{\elln\}$, and computation will still scale linearly with the landmark count $N$.
  However, it is often the case that hundreds of landmarks are being tracked and estimated, and independent computation of $\Hbarblock(x; \elln)$ for each landmark can be prohibitive.
  The restriction to finite landmark sets is additionally limiting because it makes it difficult to model how future landmarks may discovered when planning into unknown space.
  Similar in spirit to \cite{zhang2019beyond}, we would like to instead generalize the finite sum over landmarks in (\ref{eq:explicit_lambda_sum}) into an evaluation against a more general landmark \emph{distribution}.

  \subsection{A Convenient Class of Measurement Models}
    Consider the case when there exists a local parameterization of the landmark $\vc{l} \in \R^{n_l}$ such that the observation function $h(x; \ell)$ is \emph{affine} in the $\vc{l}$.
    That is, assume $h$ can be written
    \begin{equation}\label{eq:affine_model}
      h(x; \ell) = \vc{h}_0(x) + \sum_{i=1}^{n_l} l_i \vc{h}_i(x) = \sum_{i = 0} l_0 \vc{h}_i(x)
    \end{equation}
    where for simplicity we ``homogenize'' the $\vc{l}$ vector by augmenting it with a fixed zero-index element $l_0 = 1$.
    From (\ref{eq:affine_model}) it is straightforward to verify that the following conditions hold for all Lie derivatives $j$:
    \begin{equation}\label{eq:jacobian_conditions}
      \Hlj(x; \ell) = \sum_{i = 0}^{n_l} l_i \Hlji(x)
      \quad \text{ and } \quad
      \Llj(x; \ell) %= \begin{bmatrix} {}_1\Ljh(x) & {}_2\Ljh(x) & \ldots & {}_{n_l}\Ljh(x) \end{bmatrix}
                     = \Llj(x)
    \end{equation}
    These conditions, which refer to as the \textbf{Jacobian conditions}, in turn are sufficient to guarantee that the marginalizing operator $\Qn(x)$ is independent of $\elln$, and that
    \begin{equation}\label{eq:Hbarblock_affine}
      \Hbarblock(x; \elln) = \sum_{i = 0}^{n_l} l_i \Hibarblock(x).
    \end{equation}

    (\ref{eq:Hbarblock_affine}) implies that $\Hbarblock$ is affine, which allows us to write the sum-over-landmarks in (\ref{eq:explicit_lambda_sum}) as
    \begin{align}
      \sum_{n = 1}^N \sigma^2(x; \elln) \Hbarblock(x; \elln)^\T \Hbarblock(x; \elln) &= \sum_{i = 0}^{n_l} \sum_{j = 0}^{n_l} \eta_{ij}(x) \Hibarblock(x)^\T \Hibarblock[j](x)  \label{eq:affine_info_sum}
    \end{align}
    where the \emph{landmark mass coefficients}
    \begin{equation}
      \eta_{ij}(x) = \sum_{n = 1}^N \sigma^2(x; \elln) l_i l_j
    \end{equation}
    describe a visibility-weighted \emph{distribution} of landmarks over the state space $\manX$.

    Note that (\ref{eq:affine_info_sum}) only requires explicit formation of the $n_l+1$ affine coefficient matrices $\Hibarblock(x)$, rather than computing $\Hbarblock(x; \ell)$ separately for each of the $N$ landmarks.
    The $(n_l+1)^2$ distribution parameters $\eta_{ij}(x)$ then fully describe how much ``landmark'' is visible from a given state $x \in \manX$.
    If these ``mass coefficients'' were computable in constant-time, so would the total information $\Lam(x)$.

    Fortunately, there exist useful observation models for which the affine condition (\ref{eq:affine_model}) is met.
    In fact, any affine function of the relative landmark position (for example in the vehicle body frame) $\vc[b]{l} = \mx[w]{R}_b^\T ( \vc[w]{l} - \vc[w]{p} )$ has this property.
    While this category does \emph{not} include perspective camera projection, it does include orthographic projection, which has often been used as a proxy \cite{kanade1998factorization}.
    This provides a convenient approximation for visual-SLAM systems, for which observability-based planning has already demonstrated benefit \cite{zhang2018perception,falanga2018pampc,murali2019perception}.
    It is worth noting that the decomposition (\ref{eq:affine_info_sum}) is much more general than that proposed by \cite{zhang2019beyond}, applying to any visibility function $\sigma(x; \ell)$ and the full class of affine observation models.

    \begin{algorithm}
      \SetAlgoLined
      \DontPrintSemicolon
      \KwInput{Nominal state and control $x_0, u_0$}
      \KwInput{Landmark distribution $\LmDist$}
      \KwInput{Measurement timestamps $\pT = \{t_0, t_1, \ldots, t_K \}$}
      \KwInput{Prior info matrix $\S_0 = \P_0^{-1}$ over $\e_0$}
      \KwOutput{Posterior info matrix $\mx{S}_K = \P_K^{-1}$ over $\e_K$}

      \tcc{Build $\Lam_0$}
      $\vc{\eta} \gets \textsc{ComputeMassCoefficients}(x_0, \LmDist)$
      %\tcp*{get mass parameters}

      $\Lam_0 \gets \begin{bmatrix} \mx{S}_0 & \mx{0} \\ \mx{0} & \mx{I} \end{bmatrix}$
      \tcp*{init joint info over $(\e_0, \W)$}

      \For{$i,j$ in $\{0, 1, \ldots, n_l\}$} {
        $\Lam_0 \gets \Lam_0 + \eta_{ij}\, \Hibarblock[i](x_0, u_0)^\T \Hibarblock[j](x_0, u_0)$
        \tcp*{accumulate affine components}
      }

      \tcc{Propagate and marginalize}
      $\Lam_K \gets \blockM^{-\T} \Lam_0 \blockM^{-1}$
      \tcp*{see (\ref{eq:info_propogate}), can be implemented sparsely}
      %$\mx{S}_K \gets [\Lam_K]_{\e,\e} - [\Lam_K]_{\e,\W}[\Lam_K]_{\W,\W}^{-1} [\Lam_K]_{\W,\e}$
      $\mx{S}_K \gets \textsc{SchurComplement}(\Lam_K)$
      \tcp*{see (\ref{eq:W_marginalize})}

      \caption{Interval Filter Update (Affine Jacobians)}
      \label{algo:interval_update_affine}
    \end{algorithm}

  \subsection{Interpretation of the Mass Coefficients}
    Intuitively, the mass coefficients $\eta_{ij}$ represent a visibility-weighted landmark \emph{distribution}, and may be interpreted a number of ways.
    %\begin{equation}\label{mc:con}
    %  [a_{ij}(x)] \succeq 0 \qquad \text{and} \qquad  c(x) \geq 0
    %\end{equation}
    %It is not obvious when (or indeed if ever) these coefficients can be computed exactly in constant-time.
    %Nonetheless, the decomposition shown in (\ref{eq:vis_info_quad}) lends itself to a variety of interpretations.

    \textbf{An optimized implementation.}
    When computing the SIIF update against a finite set of landmarks $\{\elln\}$, (\ref{eq:affine_info_sum}) allows the affine components $\Hibarblock(x_0, u_0)$ to be computed \emph{once} for all landmarks and then directly combined.

    \textbf{Pre-computation and lookup.}
    When $\sigma$ depends on a low-dimensional component of the state space $y = g(x)$, that is that $\sigma(x; \ell) = \sigma(y; \ell)$, this $y$-space can be discretized and corresponding values of the mass coefficients can be pre-computed.
    Then computation can be approximated by a lookup table.
    For example, the authors of \cite{zhang2019beyond} do exactly this for a particular choice of $\sigma$, discretizing over a workspace in $\R^3$.

    \textbf{A (possibly learned) prior.}
    In many online robotics applications, landmarks are discovered and tracked in real-time as they enter the sensor's FoV.
    As trajectories are often planned to the edge of (or beyond) the sensing horizon, planning strictly against the \emph{currently}-estimated cloud may lead to myopic, undesirable behaviors such as ``turning-back.''
    This is because the reliance on a finite landmark set $\{\elln\}$ in (\ref{eq:explicit_lambda_sum}) affords no mechanism for \emph{anticipation} of where new landmarks may appear.
    However, if $\vc{\eta}(x)$ is interpreted as (and replaced by) a \emph{predictive} model over $\manX$, then the affine update (\ref{eq:affine_info_sum}) will naturally extend into into unknown space.

    In this paper, we leave implementation of a predictive landmark model for future work.
    For now, our experimental results focus on demonstrating the computational advantages of the affine formulation.

\section{Results}\label{s:results}
  Our approach was evaluated in simulation on a VIO-equipped quadrotor analogous to that used in the EuRoC MAV dataset \cite{burri2016euroc}.
  The 21-DoF estimator dynamics are assumed IMU-driven, with uncertain IMU biases and camera extrinsics calibration.
  The 12-DoF plant dynamics are driven by a commanded mass-normalized thrust $c$ and angular moments $\vc[b]{\tau}$.
  The full dynamics are given below.
  \begin{equation*}
    x = \begin{pmatrix} \vc[w]{p} & \vc[w]{v} & \rot[w]_b & \vc[b]{\omega} \end{pmatrix}  \qquad
    u = \begin{pmatrix} c & \vc[b]{\tau} \end{pmatrix}   \qquad
    e = \begin{pmatrix} \vctilde[w]{p} & \vctilde[w]{v} & \vctilde[w]{q}_b & \vctilde[b]{b}_a & \vctilde[b]{b}_\omega & \vctilde[c]{q}_b & \vctilde[c]{t}_b \end{pmatrix}
  \end{equation*}
  \begin{equation*}
    \begin{aligned}[t]
      &\textbf{Plant dynamics:}  \\
      &\, \ddt \vc[w]{p} = \vc[w]{v}  \\
      &\, \ddt \vc[w]{v} = \vc[w]{a} = c \rot[w]_b( \vc{e}_3 ) + \vc[w]{g} + \vc{\eta}_a  \\
      &\, \ddt \rot[w]_b = \vc[b]{\omega}  \\
      &\, \ddt \vc[b]{\omega} = \mx{J}^{-1} \big( \vc[b]{\tau} + \vc{\eta}_\tau - \skew{ \vc[b]{\omega} } \mx{J} \vc[b]{\omega} \big)  \\
      \\
      &\textbf{IMU model:}  \\
      &\, \vc[b]{\omega}_m = \vc[b]{\omega} + \vc{b}_\omega + \vc{\nu}_\omega  \\
      &\, \vc[b]{a}_m = \rot[w]_b^{-1} ( \vc[w]{a} - \vc[w]{g} ) + \vc{b}_a + \vc{\nu}_a
    \end{aligned}
    \,
    \begin{aligned}[t]
      &\textbf{Estimator dynamics:}  \\
      &\, \ddt \vchat[w]{p} = \vchat[w]{v}  \\
      &\, \ddt \vchat[w]{v} = \rothat[w]_{b} (\vc[b]{a}_m - \vchat{b}_a) + \vc[w]{g}  \\
      &\, \ddt \rothat[w]_b = \vc[b]{\omega}_m - \vchat{b}_\omega  \\
      &\textbf{Error dynamics:}  \\
      &\, \ddt \vctilde[w]{p} = \vctilde[w]{v}  \\
      &\, \ddt \vctilde[w]{v} \approx -c \mxhat[w]{R}_b \mx{E}_3 \vctilde[w]{q}_b - \mxhat[w]{R}_b \vctilde{b}_a - \vc{\nu}_a + \vc{\eta}_a   \\
      &\, \ddt \vctilde[w]{q}_b  = -\vctilde{b}_\omega - \vc{\nu}_\omega
    \end{aligned}
  \end{equation*}
  The orthographic projection model given landmark position $\vc[w]{l}(\ell)$ in $\R^3$ is
  \begin{equation}\label{eq:ortho_projection}
    h(x, \ell) = \begin{bmatrix} \mx{I}_2 & \mx{0}_{2\times1} \end{bmatrix} \Big( \mx[c]{R}_b \mx[w]{R}_b^\T (\vc[w]{l} - \vc[w]{p}) + \vc[c]{t}_b \Big)
  \end{equation}
  where the rigid-body transform $(\mx[c]{R}_b, \vc[c]{t}_b)$ describes the (uncertain) body-to-camera offset.
  In our simulation, the camera is forward-mounted and captures measurements at a rate of $1 / \Dt = 50$ Hz.

  \subsection{Validation of Lie-Taylor Approximation}

    \begin{figure}[t]
      \centering
      \includegraphics[width=\columnwidth]{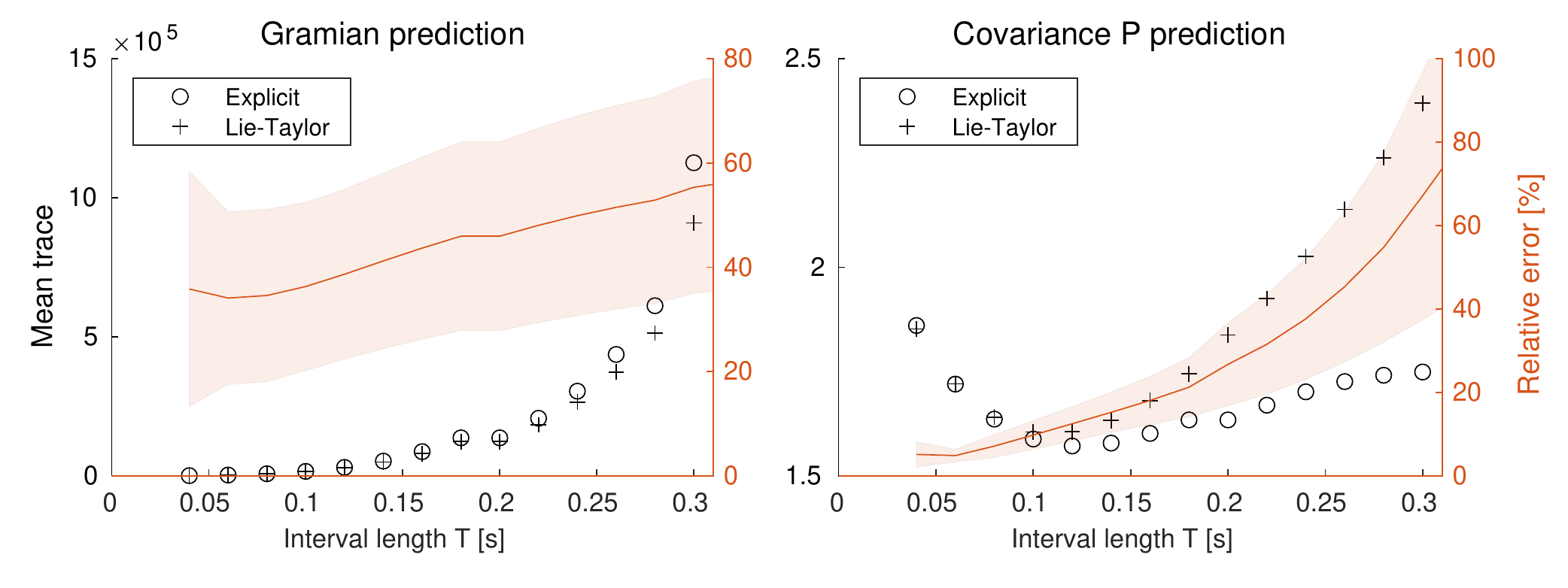}
      \vspace{-0.5cm}
      \caption{Numerical comparison between empirical and Lie-Taylor implementations for a single interval-filter update over varying interval lengths $T = K \Dt$.
        The left plot shows results for the information Gramian $[\LamZ(T)]_{\e_0,\e_0}$ (the sub-block of the information matrix pertaining to initial ego-state $\e_0$),
        and the right plot does the same for the posterior covariance $\P(T)$ given $\P(0) = 0.3 \mx{I}$.
        Unsurprisingly, both the explicit and Lie-Taylor forms of $\LamZ$ increase monotonically with $T$, as more measurements are included in the update.
        More interestingly, the \emph{relative error} between ${\LamZ}_{\text{exp}}$ and ${\LamZ}_{\text{lie}}$ is near-constant over the horizon $T$.
        \xx{Why??}
        Moving to the covariance results in the right plot, note that the traces of the estimates track well for low $T$, but ultimately diverge exponentially as should be expected.
        Nevertheless, Lie-Taylor prediction of $\P(T)$ achieves reasonable accuracy over moderate intervals.
      }
      \label{fig:sensor_approx}
    \end{figure}

    To evaluate the Lie-Taylor approximation described in Section \ref{s:siif_update}, we can compare the approximation (\ref{eq:lie_innovation}) with the explicit form (\ref{eq:explicit_innovation}).
    The results over varying interval length $T$ are plotted in Fig.~\ref{fig:sensor_approx}, and show that while error of course grows with time, satisfactory approximation is achieved for time horizons corresponding to the accumulation of 10 or more measurements in either case.
    The computational advantage of our approximation is demonstrated in Table \ref{tbl:timing}, in which computation times of the full $\Jobs$ objective are compared.

  \subsection{Value of Posterior-Covariance Objective}
    We compared the performance of the posterior-covariance objective (\ref{eq:junc_pc}) to several heuristics:
    \begin{itemize}
      \item \textbf{Maximizing landmark visibility:}
            The method labeled \ttt{max-visibility} explicitly maximizes the \emph{total visibility} along the trajectory, using the smooth visibility function given in (\ref{eq:sigma_ours})
            \begin{equation}
              \Jobs \triangleq \sum_{\kbar=1}^{\Kbar} \sum_{n = 1}^N \sigma\big(x(t_k); \ell_n\big).
            \end{equation}
      \item \textbf{Max Fisher information:}
            Following \cite{preiss2018simultaneous}, \ttt{max-gramian} maximizes the smallest eigenvalue of the information Gramian (\ie, the Fisher information)
            \begin{equation}
              \Jobs \triangleq s_1 \bigg( \sum_{\kbar=1}^{\Kbar} \mx{Y}_{\vc{v}} [\Lam_k(x_k, u_k)]_{\e_\kbar,\e_\kbar} \mx{Y}_{\vc{v}}^\T \bigg).
            \end{equation}
            Because our sensor model (\ref{eq:ortho_projection}) assumes unknown landmarks, absolute position is unobservable and therefore the position sub-matrix of $\Lam_k$ is always zero.
            Therefore we chose $\mx{Y}_{\vc{v}}$ to extract the velocity $\vctilde[w]{v}$ sub-block of $\Lam_k$.
    \end{itemize}

    Additionally, we consider the efficacy of various approximations of the posterior-covariance during planning.
    For example, \texttt{pc-exact} uses the explicit form of the SIIF update (\ref{eq:explicit_innovation}), more accurately reflecting how estimator covariance is evaluated in practice.
    The method labeled \texttt{pc-lie} applies our Lie-Taylor approximation as given in (\ref{eq:lie_innovation}) to computing $\Jobs$, and \texttt{pc-lie-cond} takes the further approximation of assuming landmarks are fully known (that is, \emph{conditioning} on rather than marginalizing out the landmark uncertainty).

    These three posterior covariance methods, alongside the heuristics, were evaluated across a batch of 50 programmatically-generated motion-planning problems.
    In each trial, a random starting state and goal position is selected, and a corresponding baseline trajectory identified by minimizing $J_c$.
    Then this trajectory is refined under the augmented objective (\ref{pblm:ct}) via IPOPT \cite{wachter2006implementation}, by replacing $J_c$ with $\Jobs$ in the objective and constraining that the conventional cost $J_c$ may not increase beyond a fixed percentage $\rflex$.
    As the tradeoff allowance $\rflex$ is increased, the refined trajectory accepts larger increases in conventional cost $J_c$ for more significant reductions in the uncertainty objective $\Jobs$.
    It is worth noting that while smooth FoV approximations (see App.~\ref{app:fov}) may be used during optimization, the output trajectories are evaluated using the explicit measurement rollout (\ref{eq:explicit_innovation}) and hard visibility indicator $\vizind(x; \ell)$.
    For our purposes here, \ttt{pc-exact} can be considered to define the Pareto front between observability and expedience.

    \begin{table}[t]
      \centering
      \caption{Averaged timing for SIIF covariance propagation on a consumer laptop over the full trajectory with $K = 7$ and $\Kbar = 11$ (includes computation of analytic gradients required for optimization).
               As can be seen, our Lie-Taylor approximation represents a significant computational speedup, and exploitation of affine sensor models as in Alg.~\ref{algo:interval_update_affine} allows for scaling to large numbers of landmarks.}
      \label{tbl:timing}
      \begin{tabular}{l@{\quad}l@{\qquad}r@{\quad}r@{\quad}r@{\quad}r}
        \toprule
        \textbf{Sensor}    & \textbf{Bundling}  & $N = 10$ & $N = 20$ & $N = 50$ & $N = 100$  \\
        \midrule
        Ortho cam          & Explicit           & 265 [ms] & 430 [ms] & 966 [ms] & 1862 [ms] \\
                           & Lie-Taylor         & \textbf{19 [ms]}  & 33 [ms]  & 77 [ms]  & 151 [ms]  \\
        Ortho cam (affine) & Explicit           & 395 [ms] & 406 [ms] & 402 [ms] & 420 [ms]  \\
                           & Lie-Taylor         & \textbf{18 [ms]}  & \textbf{17 [ms]}  & \textbf{18 [ms]}  & \textbf{17 [ms]}   \\
        \midrule
      \end{tabular}
      \vspace{-0.5cm}
    \end{table}

    \begin{figure}[h]
      \includegraphics[width=\columnwidth,trim={0.2cm, 0.4cm, 0.3cm, 0.2cm},clip]{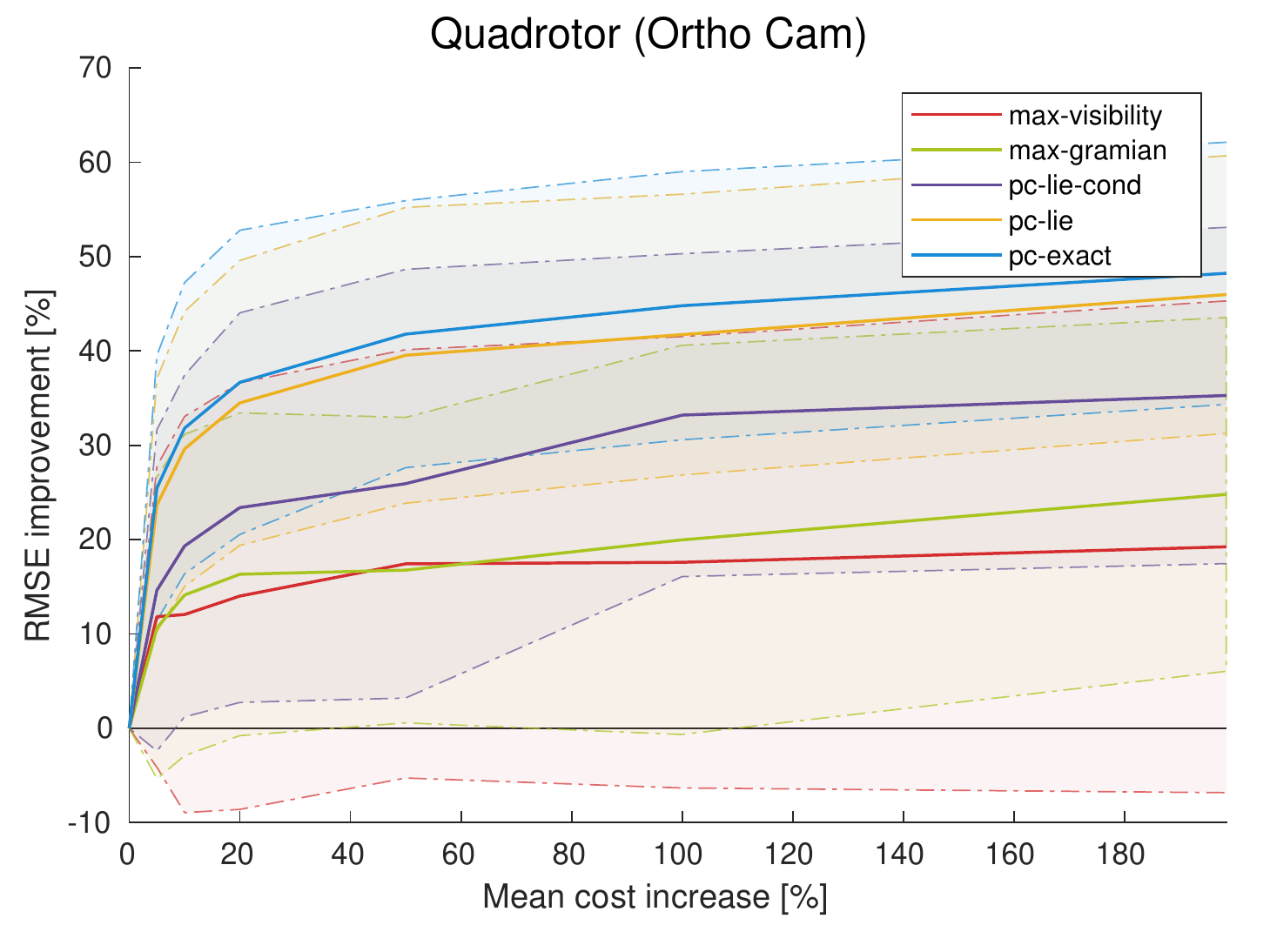}
      \caption{Each choice of $\Jobs$ defines an effective trade-off curve between trajectory cost $J_c$ and estimation improvement (larger is better).
               We plot this curve by sweeping through the weighting parameter $\rflex$ and aggregating results for a batch of random trials.
               Refinement based on heuristic objectives (see \ttt{max-viz} and \ttt{max-gramian}) or approximate sensor models (see \ttt{pc-lie-cond}) does not always yield significant estimation improvement.
               In contrast, minimization of the posterior covariance (via \ttt{pc-lie} or \ttt{pc-exact}) produces better estimation improvement for the same $J_c$ cost increase.}
      \label{fig:batch_multi}
    \end{figure}

    \begin{figure}[h]
      \includegraphics[width=\columnwidth,trim={0.0cm, 0.0cm, 0.3cm, 0.0cm},clip]{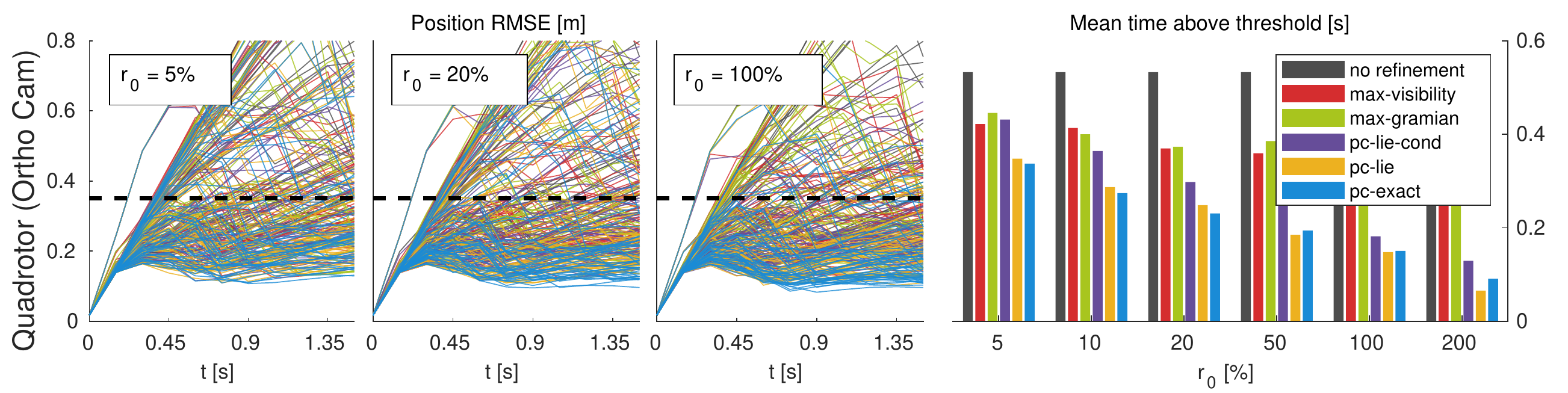}
      \caption{Evolution of the covariance traces shown for each refined trajectory, under varying settings of $\rflex$.
               For a chosen ``safety'' threshold (dashed horizontal line), histograms of expected violation time are shown on the right.
               Without any refinement (black), uncertainty can grow without bound, but refinement under posterior-covariance objectives tend to keep uncertainty bounded, even for relatively low $\rflex$.
               Other methods are significantly less effective at curbing uncertainty growth and ensuring safety.
        }
      \label{fig:cov_traces}
    \end{figure}

    Fig.~\ref{fig:batch_multi} shows that our approximation \ttt{pc-lie} is nearly as effective as \ttt{pc-exact} in guiding the solver towards well-observable trajectories, achieving significant RMSE reductions at moderate marginal cost.
    Moreover, the heuristic objectives (\ttt{max-viz} and \ttt{max-gramian}) and incorrect sensor model (\ttt{pc-lie-cond}) are far less effective, often pushing the solver towards irrelevant and unhelpful solutions.

    A central goal of observability-based planning is to avoid instances where uncertainty grows to dangerous or catastrophic levels.
    Fig.~\ref{fig:cov_traces} plots position uncertainties along a large number of simulated trajectories for the two systems.
    Unrefined trajectories can develop large uncertainty, often well beyond safe levels, and heuristic refinement methods do not ensure good performance in all instances (even if they do well on average).
    In contrast, refinement via the \ttt{pc-lie} objective effectively moderates uncertainty growth across a variety of conditions, ensuring safety.

    Note that our computational complexity results in Table~\ref{tbl:timing} demonstrate that \ttt{pc-lie} is indeed far less expensive than \ttt{pc-exact}, while being nearly as effective in guiding the optimizer towards well-observable trajectories.
    This further validates our Lie-Taylor approximation and demonstrates its applicability for online trajectory generation.

\section{Conclusions}
  Our results indicate that significant estimation improvement can be achieved by explicitly considering observability during trajectory generation.
  While posterior-covariance minimization is not novel in itself, we address several computational challenges that arise in the case of landmark-based estimators (i.e.,\ visual SLAM).
  In doing so, we reduce algorithmic complexity from quadratic to near constant in the number of landmarks $N$, opening the door to online, real-time observability-aware planning.
  Furthermore, we identify a natural mechanism allowing for generic (and non-finite) landmark \emph{distributions}, which ultimately enables extension of estimator performance prediction into unknown space.

  In ongoing work, we are developing a real-time implementation capable of online motion planning for a VIO-enabled quadrotor.
  Furthermore, we hope to further explore the idea of learned landmark distributions.

  Additionally, it is likely that orthographic projection will not ultimately provide an optimal analog for real-world cameras, which are usually modeled under perspective projection.
  Identifying a more accurate affine approximation of perspective projection is of direct interest, but for now is left as the subject of future work.

\section{Acknowledgment}
  This work was supported by the Education Office at the Charles Stark Draper Laboratory, Inc.\ and by ARL DCIST under Cooperative Agreement Number W911NF-17-2-0181.
\bibliographystyle{spmpsci}
\bibliography{IEEEabrv,ref}

\appendix

\section{Modeling Field-of-View}\label{app:fov}
  Because landmarks are often distributed non-uniformly, it is important that generated trajectories point sensors towards informative regions of the environment.
  For the purposes of continuous observability-aware optimization, we (similar to \cite{zhang2019beyond,murali2019perception}) seek a function $\sigma(x; \ell) \in [0,1]$ that \emph{smoothly} interpolates the ``hard'' visibility indicator $\vizind(x; \ell) \in \{0, 1\}$.

  As observed in Alg.~\ref{algo:interval_update} and \ref{algo:interval_update_affine}, for our purposes we actually use the squared visibility $\sigma^2(x; \ell)$.
  One choice well-suited for a pinhole camera model is
  \begin{equation}\label{eq:sigma_ours}
    \sigma^2(x, \ell) =
      \begin{cases}
        \frac{1}{2} \big( \cos a\theta + 1 \big) & |\theta| < \theta_{\text{max}}  \\
        0 & \text{else}
      \end{cases}
  \end{equation}
  where
  \begin{equation}
    \theta = \cos^{-1}\left( \frac{ \hat{\vc{e}}^\T \vc[c]{l} }{\vc[c]{l}^\T \vc[c]{l}} \right) \qquad \in [0, \pi]
  \end{equation}
  is the angle between the optical axis $\hat{\vc{e}}$ and the landmark vector $\vc[c]{l}$ in the camera frame.
  The scaling parameter can be chosen $a = \pi / \theta_{\text{max}}$, ensuring that $\sigma^2$ is continuous and differentiable.
  In practice $\theta_\text{max}$ can be chosen to exactly match the (assumed conical) FoV of the sensor, or taken slightly larger to improve convergence of the optimization.

  \begin{figure}[t]
    \centering
    \includegraphics[width=0.99\columnwidth]{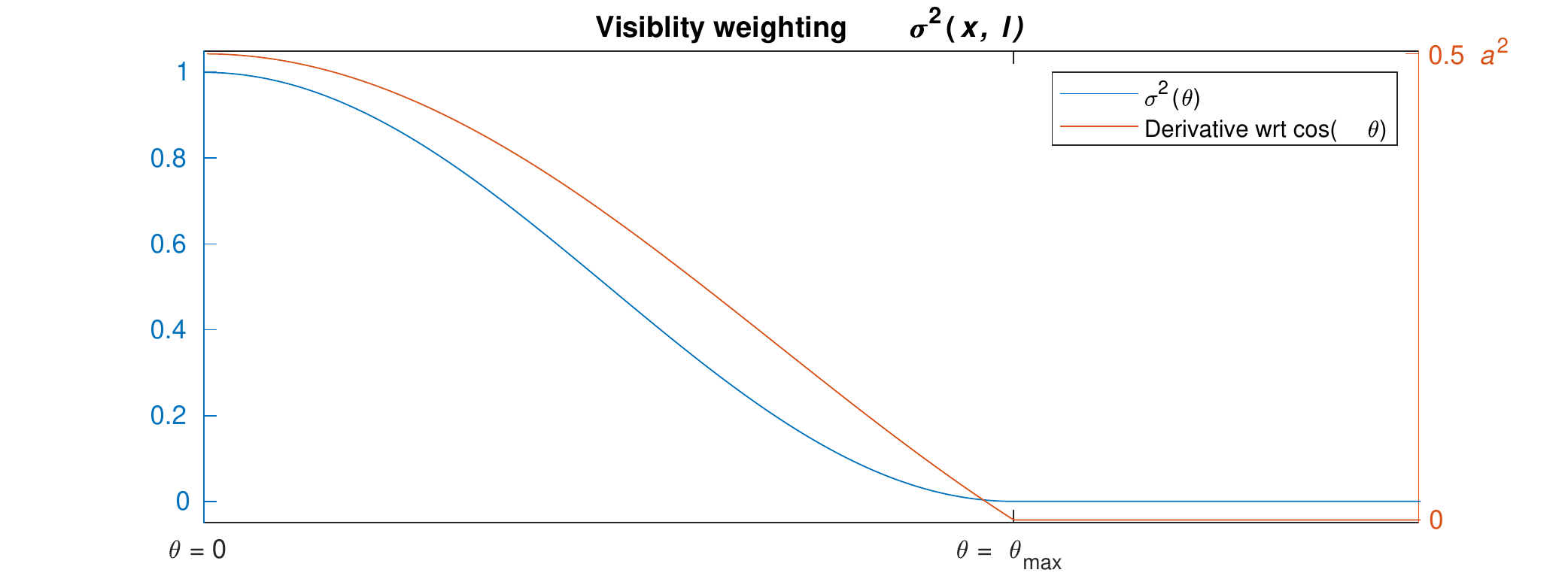}
    \caption{Our visibility weighting function (\ref{eq:sigma_ours}) plotted against $\theta$ [blue].
             The derivative $\frac{ \partial \sigma^2(\cos \theta) }{ \partial \cos \theta }$ is plotted as a function of $\theta$ in red, and both are clearly continuous (and bounded) over the entire domain $[0, \pi]$.
             }
    \label{fig:sigma_viz}
  \end{figure}

  It may appear problematic that the inverse cosine $\cos^{-1}(y)$ is not differentiable in its argument at $y = \cos \theta = 1 \iff \theta = 0$ (and its derivative in the open interval $\theta \in (0, \pi]$ is unbounded).
  Fortunately, the composite function $\sigma^2(y)$ \emph{is} differentiable in $y$ for all $\theta \in [0, \pi]$ with bounded derivative, as shown in Fig.~\ref{fig:sigma_viz}.

\section{Computing Gradients through Marginalization}\label{app:diff_marg}
  As pointed out by Mourikis \etal \cite{mourikis2007multi}, computing a left-nullspace matrix $\Qn$ of $\Lnblock$ satisfying (\ref{eq:null_Qn}) can be accomplished via a partial SVD.
  \begin{equation}
     \Lnblock = \mx{U} \begin{bmatrix} \mx{\Sigma}_1 \\ \mx{0}_{r \times n} \end{bmatrix} \mx{V}^\T
     \quad \implies \quad
     \Qn      = \mx{U} \begin{bmatrix} \mx{0}_{n \times r} \\ \mx{I}_r \end{bmatrix}
  \end{equation}
  Then the linearized residual (\ref{eq:ltv_batch_H}) can be made independent of landmark error $\lmntilde$ by application of $\Qn^\T$.
  However, in a continuous optimization framework, we need to be able to compute gradients through this marginalization process.

  While the SVD is differentiable in general \cite{townsend2016differentiating}, we identify a simpler form of $\diff \Qn$.
  Letting $\Lnblock$ have dimension $m \times n$ with $m > n$, then $\Qn$ will have dimension $m \times (m-n)$.
  Letting $r = m - n$, then a suitable $\diff \Qn$ must also be $m \times r$ and satisfy differentiated forms of the constraints (\ref{eq:null_Qn}).
  \begin{align}
    \diff \Qn^\T \Lnblock + \Qn^\T \diff \Lnblock &= \mx{0}_{r \times n}  \\
    \diff \Qn^\T \Qn + \Qn^\T \diff \Qn &= \mx{0}_{r \times r}
  \end{align}
  It is straightforward to verify that these conditions will be satisfied by any $\diff \Qn$ such that
  \begin{align}
    \diff \Qn &= \mx{U} \begin{bmatrix} \mx{Z}_1 \\ \mx{Z}_2 \end{bmatrix}  \\
    \mx{Z}_1 &= -\mx{\Sigma}_1^{-1} \mx{V}^\T \diff \mx{H}_{\ell}^\T \Qn  \\
    \mx{Z}_2 &= -\mx{Z}_2^\T \quad \text{(skew-symmetric)}
  \end{align}
  where the inverse of diagonal $\mx{\Sigma}_1$ is trivial to compute, and $\mx{Z}_2$ can conveniently be taken as $\mx{0}_{n \times n}$.
  In practice, this recipe is inexpensive to compute and straightforward to implement.

\end{document}